\documentclass[a4paper]{spie}  

\usepackage{amsmath,amsfonts,amssymb}
\usepackage{graphicx}
\usepackage[colorlinks=true, allcolors=blue]{hyperref}

\title{Toward explainable AI approaches for breast imaging: adapting foundation models to diverse populations}

\author[a, b]{Guilherme J. Cavalcante}
\author[b]{José Gabriel A. Moreira}
\author[a, b]{Gabriel A.B. do Nascimento}
\author[a]{Vincent Dong}
\author[a]{Alex Nguyen}
\author[b]{Thaís G. do Rêgo}
\author[b]{Yuri Malheiros}
\author[c]{Telmo M. Silva Filho}
\author[a]{Carla R. Zeballos Torrez}
\author[a]{James C. Gee}
\author[a]{Anne Marie McCarthy}
\author[a]{Andrew D. A. Maidment}
\author[a]{Bruno Barufaldi}

\affil[a]{University of Pennsylvania, Philadelphia, PA, US}
\affil[b]{Federal University of Paraíba, João Pessoa, PB, Brazil}
\affil[c]{University of Bristol, Bristol, UK}

\begin{document} 
\maketitle

\begin{abstract}
Foundation models hold promise for specialized medical imaging tasks, though their effectiveness in breast imaging remains underexplored. This study leverages BiomedCLIP as a foundation model to address challenges in model generalizations. BiomedCLIP was adapted for automated BI-RADS breast density classification using multi-modality mammographic data (synthesized 2D images, digital mammography, and digital breast tomosynthesis). Using 96,995 images, we compared single-modality (s2D only) and multi-modality training approaches, addressing class imbalance through weighted contrastive learning. Both approaches achieved similar accuracy (multi-modality: 0.74, single-modality: 0.73), with the multi-modality model offering broader applicability across different imaging modalities and higher AUC values consistently above 0.84 across BI-RADS categories. External validation on the RSNA and EMBED datasets showed strong generalization capabilities (AUC range: 0.80–0.93). GradCAM visualizations confirmed consistent and clinically relevant attention patterns, highlighting the models' interpretability and robustness. This research underscores the potential of foundation models for breast imaging applications, paving the way for future extensions for diagnostic tasks.

\end{abstract}


\section{INTRODUCTION}
\label{sec:intro}


Biomedical foundation models can reduce the reliance on large annotated datasets by learning from broad image-text pairs across medical domains. BiomedCLIP~\cite{biomedclip}, an adaptation of Contrastive Language-Image Pre-training (CLIP)~\cite{clip}, is a foundation model that has been pre-trained on 15 million image-text pairs extracted from PubMed Central articles. It is trained using contrastive learning to encode joint representations of image and text, and has demonstrated improved performance in radiology and pathology tasks including chest X-ray interpretation, medical image retrieval, and zero-shot medical image classification. However, while these models excel at general downstream tasks, their performance in specialized domains without task-specific fine-tuning remains limited~\cite{molina2024comparison}. Recent studies have shown that BiomedCLIP struggles with mammographic density interpretation in its pre-trained state, achieving suboptimal results without domain-specific adaptation~\cite{molina2024comparison}. This limitation impedes broader usage and clinical adoption; Breast cancer screening presents unique challenges (e.g., tissue classification and lesion detection), requiring specialized domain adaptation.

Breast density impacts cancer screening by masking tumors and reducing the sensitivity of 2D digital mammography (DM)\cite{carney2003}. It is subjectively classified into four BI-RADS categories (A–D), with significant inter-reader variability\cite{sprague2016variation}, highlighting the need for automated assessment. Digital breast tomosynthesis (DBT) addresses detection limitations by reconstructing 3D breast images~\cite{friedewald2014}. Synthesized 2D (s2D) images from DBT projections offer similar detection performance to DM while simplifying data for AI development~\cite{skaane2014,zuley2014}.

We adapted BiomedCLIP~\cite{biomedclip} for downstream tasks in breast imaging, focusing on BI-RADS density classification. The model is fine-tuned using multi-modality data (DM, DBT central slices, and s2D images) to improve cross-modality generalization, and weighted contrastive learning is used to address class imbalance. We compared single-modality (s2D only, single site) and multi-modality (s2D, DM, and DBT, multi-site) model performance. Zero-shot validation is performed on public datasets.

\section{MATERIALS AND METHODS}

\subsection{Dataset}

This retrospective study used imaging data from two breast cancer screening sites within the University of Pennsylvania Health System (2011-2022). Each screening study consists of two standard views (craniocaudal and mediolateral oblique) per breast, with additional views acquired when required. The data (Table \ref{tab:demographic_characteristics}) with annotations and reports by radiologists according to BI-RADS density categories (A-D). 

Two experiments were conducted using (1) s2D only (14,911 images) and (2) multi-modality (96,995 images combining s2D, DM, and DBT central slices). The multi-modality dataset included images from Hologic and Siemens DBT systems, with 10,060 Siemens images (10.4\%) and the remaining 86,935 images (89.6\%) from Hologic. To address BI-RADS density class imbalance, we implemented a systematic approach to determine optimal class distributions. Starting with the minority class (D density), we used all available D density images (3,000) as the baseline and conducted iterative experiments with different undersampling ratios for the remaining classes. The final distribution (4,000 A, 6,000 B, 6,000 C, and 3,000 D images) was selected based on validation performance, balancing the need for sufficient representation of each class while maximizing overall model accuracy. This distribution maintained 6,253 Siemens images (32.9\%) and 12,747 Hologic images (67.1\%). The modality distribution was approximately 50\% s2D, 25\% DM, and 25\% DBT, with greater number of s2Ds reflecting the prospective use of DBT+s2D mode in screening. Similarly, for the s2D only model, the data was undersampled to maintain similar proportions; all available D density images were used (minority class), resulting in 920 A, 1,380 B, 1,380 C, and 644 D images (4,324 total), all from Hologic systems. For external validation, the multi-modality model was validated on the publicly available RSNA (29,470 images) and EMBED (22,937 images) datasets in a zero-shot classification setting.

\begin{table}[h]
    \centering
    \caption{Demographic characteristics of the study population stratified by imaging modality.}
    \label{tab:demographic_characteristics}
    \begin{tabular}{p{5.2cm}p{2.5cm}p{2.5cm}p{2.5cm}p{2cm}}
    \hline
    \textbf{Demographics} & \textbf{s2D} & \textbf{DBT} & \textbf{DM} & \textbf{\textit{P} value*} \\
    \hline
    Studies, \textit{count} & 3,184 & 5,813 & 13,137 & \\
    Age, \textit{mean years (±SD)} & 54.93 (9.70) & 56.93 (9.87) & 56.94 (10.23) & $<$1e-20 \\
    Race, \textit{count (\%)} & & & & $<$0.001\\
    \quad Black & 1,550 (48.6) & 2,861 (49.1) & 6,131 (46.6) & \\
    \quad White & 1,452 (45.6) & 2,655 (45.6) & 6,142 (46.7) & \\
    \quad Other & 182 (5.7) & 297 (5.1) & 864 (6.5) & \\
    BI-RADS Density, \textit{count (\%)} & & & & 0.029\\
    \quad A & 362 (11.3) & 714 (12.2) & 1,504 (11.4) & \\
    \quad B & 1,709 (53.6) & 3,117 (53.6) & 6,974 (53.0) & \\
    \quad C & 971 (30.4) & 1,803 (31.0) & 4,153 (31.6) & \\
    \quad D & 142 (4.4) & 179 (3.0) & 506 (4.8) &\\
    \hline
    \multicolumn{5}{l}{*Significant differences across modalities (e.g., age: p$<$1e-20), but effect sizes are small (Cramer’s V $<$0.03;}\\
    \multicolumn{5}{l}{age difference $\sim$2 years), indicating no practically meaningful imbalance. Pairwise comparisons between} \\ 
    \multicolumn{5}{l}{DBT and DM also showed minimal imbalance (Cramer’s V $<$0.08) with no differences in age (p=0.72).} \\
    \end{tabular}
\end{table}

\subsection{Model Training}
The pipeline used to develop and train the models is described in the following steps:

\textbf{Image preprocessing: }Non-anatomical artifacts such as burned-in annotations and scanner-generated text overlays were removed, and breast regions were isolated by cropping and excluding background and compression artifacts. The resulting images were then resized to a resolution of 224 × 224 pixels, and normalized to standardize contrast and brightness across different imaging protocols.

\textbf{Report simplification: }Radiologists provided structural reports of each patient. The textual elements were simplified to the description of the BI-RADS density categories: ``fatty or almost entirely fatty breasts", ``scattered areas of fibroglandular density",``heterogeneously dense breasts", and ``extremely dense breasts".

\textbf{Foundation model: }Our approach fine-tunes BiomedCLIP by adapting its pre-trained vision and text encoders to breast density classification. During training, ground truth image-text pairs with matching classes formed positive pairs, while mismatched pairs from different classes served as negatives. Both encoders project their outputs to a shared embedding space where the model learns to maximize cosine similarity for positive pairs and minimize it for negative pairs via a weighted contrastive loss. As a result of this training, the model employs a dual-encoder architecture that processes both mammographic images and textual descriptions of breast density categories. The vision encoder extracts visual features relevant to breast density assessment, while the text encoder processes fixed textual prompts describing BI-RADS density categories.

\textbf{Data stratification: }To address the class imbalance present in our datasets, we implemented a weighted contrastive learning approach that gives higher importance to underrepresented classes during training. Class weights were computed as the inverse frequency of each class, normalized to maintain balanced training. This approach ensures that the model learns to recognize patterns in all density categories, even those with fewer training examples. To prevent data leakage and ensure proper evaluation, we implemented a stratified group splitting strategy that respects the patient-level data distribution. This approach ensures that images from the same patient are not divided between training and validation sets, with longitudinal data (multiple examinations from the same patient over time) assigned exclusively to training sets and single-examination cases distributed across folds while maintaining class balance.

\textbf{Cross-fold validation: }We employed a stratified 5-fold cross-validation strategy to maximize the available training data while providing robust estimates of model performance across different data subsets. This approach helps identify potential overfitting while maintaining representative class distributions across all folds, ensuring reliable performance assessment despite the severe class imbalance present in the dataset.

\section{RESULTS AND DISCUSSION}


We evaluated our foundation model approach across two main experiments (single-modality and multi-modality) and multiple external validation datasets to assess generalization capabilities. External validation was performed using the multi-modality model on the RSNA and EMBED datasets. Preliminary performance was quantified using classification metrics such as AUC and accuracy. Table~\ref{tab:performance_metrics} presents the results for each BI-RADS density class across all experiments and datasets.

\begin{table}[h]
    \centering
    \caption{Performance metrics for each BI-RADS density class (A-D) across experiments.}
    \label{tab:performance_metrics}
    \small
    \begin{tabular}{|l|cccc|cccc|cccc|cccc|}
    \hline
    & \multicolumn{4}{c|}{\textbf{Penn (s2D only)}} & \multicolumn{4}{c|}{\textbf{Penn (s2D,DM,DBT)}} & \multicolumn{4}{c|}{\textbf{RSNA (DM only)}} & \multicolumn{4}{c|}{\textbf{EMBED (DM only)}} \\
    \cline{2-17}
    & A & B & C & D & A & B & C & D & A & B & C & D & A & B & C & D \\
    \hline
    AUC & 0.90 & 0.86 & 0.85 & 0.93 & 0.91 & 0.86 & 0.84 & 0.94 & 0.94 & 0.81 & 0.88 & 0.91 & 0.92 & 0.82 & 0.84 & 0.91 \\
    Acc & 0.79 & 0.66 & 0.74 & 0.73 & 0.81 & 0.66 & 0.72 & 0.75 & 0.78 & 0.65 & 0.64 & 0.60 & 0.63 & 0.64 & 0.64 & 0.71 \\
    \hline
    \end{tabular}
\end{table}

Balanced performance across BI-RADS categories was observed, with the multi-modality approach showing slightly improved accuracy compared to s2D-only, despite having a marginally lower AUC. The performance between s2D-only and multi-modality models suggests that the multi-modality approach is preferable due to ability to handle diverse imaging settings and systems. Errors in misclassifications occurred predominantly between adjacent classes (e.g., A→B, B→C, C→D), particularly in cases where density classification is more subjective. Importantly, RSNA and EMBED contains imaging variations (e.g., paddle marks, focal spot, etc.) not included in the training; the model is insensitive to these variations in zero-shot classification, demonstrating ability to learn fundamental tissue information rather than relying on dataset-specific features.

\begin{figure}[h]
    \centering
    \includegraphics[width=0.8\linewidth]{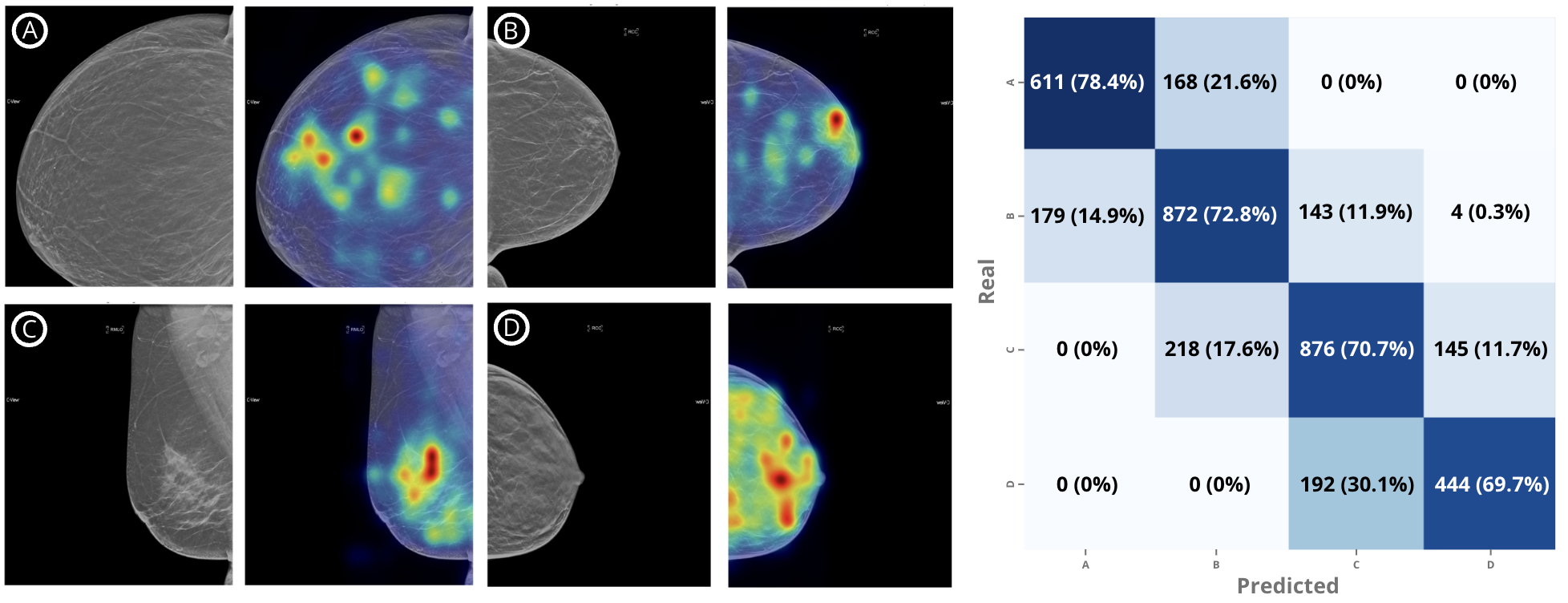}
    \caption{Grad-CAM visualizations across BI-RADS categories A–D (left) and confusion matrix on the validation set (right), using data from a single site (s2D only). \textbf{Note:} Warmer colors indicate higher model attention.}
    \label{fig:gradcam}
\end{figure}

\begin{figure}[h]
    \centering
    \includegraphics[width=\linewidth]{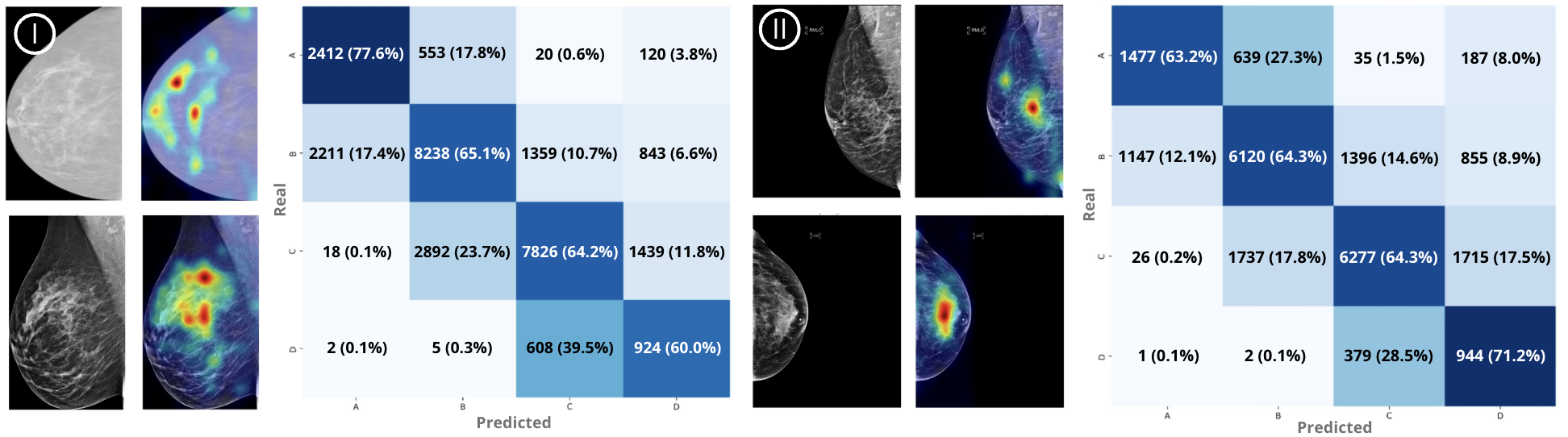}
    \caption{Grad-CAM visualizations and confusion matrices for two external datasets. Left panel (I): two RSNA examples, each showing the original image (left) and the corresponding Grad-CAM overlay (right), along with the confusion matrix. Right panel (II): two EMBED examples with the same layout, and the associated confusion matrix shown on the right.}
    \label{fig:gradcam_external}
\end{figure}

We use GradCAM \cite{gradcam} to generate saliency maps (Figure~\ref{fig:gradcam})  to show the attention patterns of the model and to identify the features that differentiate the four breast density grades.   In fatty or almost entirely fatty breasts (Fig.~\ref{fig:gradcam}A), the model's attention is diffusely distributed across the entire breast area. In contrast, for scattered (Fig.~\ref{fig:gradcam}B), heterogeneously dense (Fig.~\ref{fig:gradcam}C), and extremely dense breasts (Fig.~\ref{fig:gradcam}D), the model's attention is concentrated in regions with higher amounts of fibroglandular tissue. Zero-shot classifications on external dataset shows that the model maintains consistent attention even on unseen data (Figure~\ref{fig:gradcam_external}).

\begin{figure}[!ht]
    \centering
    \includegraphics[width=\linewidth]{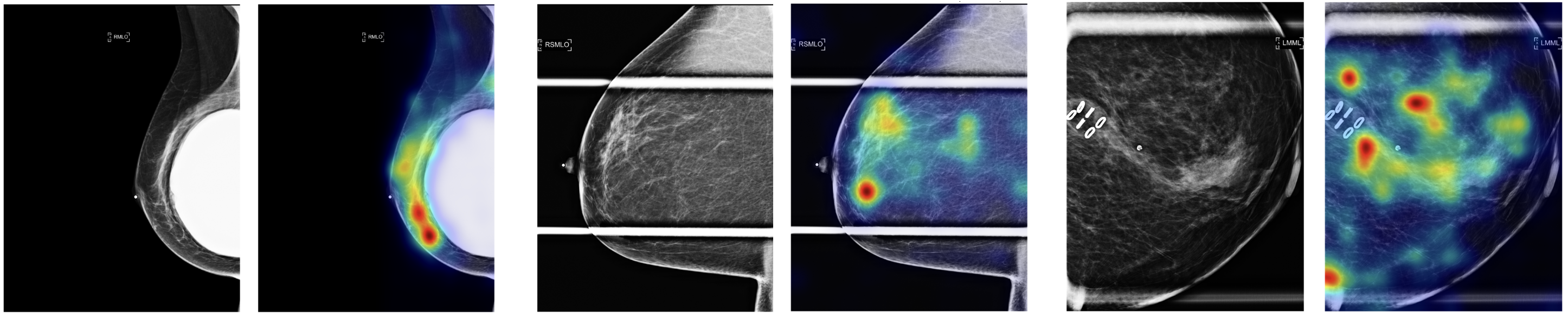}
    \caption{GradCAM visualizations demonstrating model robustness to imaging variations (paddles, annotations, implants) while maintaining focus on breast tissue density patterns.}
    \label{fig:gradcam_robustness}
\end{figure}

Model robustness was visually evaluated on external datasets with significant variations in imaging protocols. Figure~\ref{fig:gradcam_robustness} demonstrates the model's ability to maintain accurate predictions while focusing attention on anatomically relevant tissues, even in the presence of compression paddles, clips, and implants; The model focus on density patterns and fundamental characteristics, rather than relying on dataset-specific features.

\section{CONCLUSIONS}

In this work, we demonstrate that a foundation model, adapted from BiomedCLIP and fine-tuned for breast imaging, can effectively classify BI-RADS breast density categories using both single- and multi-modality mammographic data. Importantly, the models are also multi-modal, incorporating features extracted from both images and text to enhance prediction accuracy and interpretability. The model achieves balanced performance across all density classes, generalizes well to external datasets, and remains robust to imaging artifacts and protocol variations. GradCAM visualizations confirm that the model focuses on clinically relevant regions, reinforcing its interpretability and potential for clinical integration. Future work will explore the extension of this framework to additional breast imaging tasks, such as lesion detection, image retrieval, and broader diagnostic applications, further advancing the role of explainable foundation models in breast imaging.

\textbf{Disclosure}: This work has not been submitted for review in any other publication.

\bibliography{report} 
\bibliographystyle{spiebib} 
\end{document}